\documentclass[10pt,twocolumn,letterpaper]{article}

\usepackage{cvpr}              % To produce the CAMERA-READY version
\usepackage[dvipsnames]{xcolor}

\definecolor{cvprblue}{rgb}{0.21,0.49,0.74}
\usepackage[pagebackref,breaklinks,colorlinks,citecolor=cvprblue]{hyperref}
\usepackage{orcidlink}
\def\paperID{*****} % *** Enter the Paper ID here
\def\confName{CVPR}
\def\confYear{2024}

\title{Multi-Stage Residual-Aware Unsupervised Deep Learning Framework for Consistent Ultrasound Strain Elastography}

\author{Shourov Joarder \orcidlink{0009-0005-7105-4035}, Tushar Talukder Showrav \orcidlink{0009-0004-4457-701X}, Md. Kamrul Hasan \orcidlink{0000-0002-4816-2725}\\
Bangladesh University of Engineering and Technology, Dhaka, Bangladesh\\
{\tt\small 1906156@eee.buet.ac.bd, 1706087@eee.buet.ac.bd, khasan@eee.buet.ac.bd}
\vspace{-.40cm} % <--- ADD THIS HERE
}

\begin{document}
\maketitle
\vspace{-3.0cm}
\begin{abstract}
Ultrasound Strain Elastography (USE) is a powerful non-invasive imaging technique for assessing tissue mechanical properties, offering crucial diagnostic value across diverse clinical applications. However, its clinical application remains limited by tissue decorrelation noise, scarcity of ground truth, and inconsistent strain estimation under different deformation conditions. Overcoming these barriers, we propose MUSSE-Net, a residual-aware, multi-stage unsupervised sequential deep learning framework designed for robust and consistent strain estimation. At its backbone lies our proposed USSE-Net, an end-to-end multi-stream encoder–decoder architecture that parallelly processes pre- and post-deformation RF sequences to estimate displacement fields and axial strains. The novel architecture incorporates Context-Aware Complementary Feature Fusion (CACFF)-based encoder with Tri-Cross Attention (TCA) bottleneck with a Cross-Attentive Fusion (CAF)-based sequential decoder. To ensure temporal coherence and strain stability across varying deformation levels, this architecture leverages a tailored consistency loss. Finally, with the MUSSE-Net framework, a secondary residual refinement stage further enhances accuracy and suppresses noise. Extensive validation on simulation, in vivo, and private clinical datasets from Bangladesh University of Engineering and Technology (BUET) medical center, demonstrates MUSSE-Net’s outperformed existing unsupervised approaches. On MUSSE-Net achieves state-of-the-art performance with a target SNR of 24.54, background SNR of 132.76, CNR of 59.81, and elastographic SNR of 9.73 on simulation data. In particular, on the BUET dataset, MUSSE-Net produces strain maps with enhanced lesion-to-background contrast and significant noise suppression yielding clinically interpretable strain patterns. 
%These findings highlight MUSSE-Net’s potential as a reliable, accurate, and scalable framework for real-time USE, paving the way toward broader clinical adoption of deep learning-based elastography.
\end{abstract}
    
\section{Introduction}
{Q}uasi-static Ultrasound Strain Elastography (USE) is a well-established, noninvasive physics-informed imaging modality that leverages tissue deformation under mechanical stress to estimate strain and infer viscoelastic properties. Its portability, cost-effectiveness \cite{b3}, and compatibility with conventional ultrasound systems have enabled widespread clinical adoption across a range of applications, including breast lesion characterization \cite{b3}, liver fibrosis staging \cite{b2}, prostate and thyroid evaluation \cite{b6,b4}, renal and vascular assessment \cite{b7}, and musculoskeletal diagnostics \cite{b8}. In quasi-static USE, axial compression is manually applied via a handheld transducer, and radio-frequency (RF) echo frames are acquired before and after deformation. Strain estimation is performed by tracking speckle displacement between pre- and post-compression frames, followed by spatial differentiation or direct inversion techniques such as least-squares strain estimation (LSQSE) to generate high-resolution strain maps.

% %%___HIGHLIGHTS___%%
% \begin{table*}[!t]
% \arrayrulecolor{subsectioncolor}
% \setlength{\arrayrulewidth}{1pt}
% {\sffamily\bfseries\begin{tabular}{lp{6.75in}}\hline
% \rowcolor{abstractbg}\multicolumn{2}{l}{\color{subsectioncolor}{\itshape
% Highlights}{\Huge\strut}}\\
% \rowcolor{abstractbg}$\bullet$ & MUSSE-Net, the first residual-aware multi-stage unsupervised sequential deep learning framework for ultrasound strain elastography, is proposed for robust and progressive refinement of displacement and strain estimation. \\

% \rowcolor{abstractbg}$\bullet$ & A sequential unsupervised deep learning approach is implemented to achieve high-quality, temporally consistent displacement tracking and strain estimation. \\

% \rowcolor{abstractbg}$\bullet$ & Compared to other state-of-the-art deep learning methods, both USSE-Net and MUSSE-Net demonstrate superior performance on simulated and \textit{in vivo} datasets. \\

% \end{tabular}}
% \setlength{\arrayrulewidth}{0.4pt}
% \arrayrulecolor{black}
% \end{table*}

Several studies \cite{b9, b10} have examined classical window-based and energy-minimization techniques \cite{b21, b10_} for displacement tracking in USE, though these non–deep learning methods are often susceptible to signal noise and decorrelation \cite{b11}. To overcome these limitations, Hussain et al. \cite{b11_} proposed a neighborhood-average strategy. While traditional speckle tracking algorithms have long supported strain estimation, deep learning–based approaches have recently gained traction, notably in \cite{b11}, driven by advances in computer vision. Models such as FlowNet \cite{b13}, FlowNet 2.0 \cite{b14}, PWCNet \cite{b15}, and RAFT \cite{b16}—originally developed for optical flow estimation—have shown promising results in USE due to the similarity in displacement tracking tasks \cite{b17}. Bo Peng \emph{et al.} \cite{b18} were among the first to adapt CNN-based speckle tracking to USE using FlowNet 2.0 retrained on ultrasound data, which yielded improvements but suffered from low SNR, poor strain image quality, and limited accuracy at low strain levels. To address these limitations, Kibria \emph{et al.} \cite{b19} proposed GLUENet, which refines displacement fields via Global Time-Delay Estimation (GLUE) \cite{b20_}, although its performance depends heavily on the quality of initial estimates. A modified version of FlowNet 2.0 was later trained for sub-pixel accuracy in \cite{b20}, but its high computational cost (nearly 160 million parameters) and lack of temporal consistency remain significant drawbacks.

In \cite{b17}, following their previous work, Bo Peng \emph{et al.} applied another widely used optical flow model PWC-Net \cite{b15} to breast ultrasound speckle tracking. Although the proposed PWC-Net based method outperformed other CNN-based approaches, its tracking accuracy remained inferior to traditional coupled tracking methods when tested on both simulated \textit{phantom} and \textit{in-vivo} datasets.
Motivated by the success of earlier pyramidal networks in this domain, Tehrani \emph{et al.} proposed two enhanced versions of PWC-Net (i.e., MPWCNet and RF-MPWCNet) to improve displacement tracking performance \cite{b31}. But these methods are not optimal for estimating displacements directly from RF data, which contain high-frequency components unlike natural images. Addressing this, some approaches incorporated B-mode images derived from RF data, as their frequency characteristics more closely resemble those of natural images \cite{b17}. Nonetheless, B-mode images lack the high-frequency information crucial to RF signals. To mitigate this limitation, studies such as \cite{b31}, \cite{b1}, and \cite{b24} incorporated both RF and B-mode data during training to preserve the full information spectrum.

A persistent challenge in developing deep learning models for USE is the scarcity of annotated ultrasound datasets \cite{b25}. To mitigate this, recent approaches such as MPWCNet \cite{b19} and RFMPWCNet \cite{b31} have employed transfer learning, pretraining on large-scale natural image datasets before fine-tuning on ultrasound data. However, the substantial domain gap between natural and ultrasound images limits the effectiveness of this strategy \cite{b25}. In response, semi-supervised and unsupervised learning methods have gained momentum. An unsupervised CNN with a simple encoder–decoder architecture and skip connections was proposed in \cite{b26}, trained using unsupervised loss functions. Building on this, Tehrani \emph{et al.} introduced MPWCNet++ \cite{b27}, supporting both supervised and unsupervised paradigms. In \cite{b11}, an updated Global Correlation Module (GoCor), originally introduced in \cite{b28}, was integrated into an unsupervised framework for speckle tracking. Wei \emph{et al.} \cite{b29} further advanced the field with a MaskFlowNet-based unsupervised model that outperformed MPWCNet, RFMPWCNet, and MPWCNet++ in SNR, contrast-to-noise ratio (CNR), and normalized root mean square error (NRMSE). Despite these gains, simple encoder–decoder architectures struggle to deliver temporally consistent strain estimates. ReUSENet \cite{b30} addresses this by incorporating ConvLSTM units within a recurrent neural network (RNN) framework, improving temporal coherence in displacement and strain estimation. However, due to the absence of explicit tracking layers and reliance on a basic encoder–decoder backbone, ReUSENet still produces strain images with notable background noise.

In this work, we propose Residual-Aware Multi-Stage Unsupervised Sequential Strain Estimation Network (MUSSE-Net), a novel deep learning framework tailored for high-fidelity strain estimation from raw ultrasound RF data. MUSSE-Net operates in a fully unsupervised manner, directly addressing the challenge of limited annotated ultrasound data. By integrating a sequential decoder with ConvLSTM units, it effectively models temporal dependencies across ultrasound frames, an aspect often neglected in existing architectures. The network builds upon our foundational multi-stream encoder–decoder design, USSE-Net, which incorporates key innovations to enhance strain estimation. Specifically, USSE-Net replaces conventional encoders with a multi-stream encoder featuring Context-Aware Complementary Feature Fusion (CACFF) and a Tri-Cross Attention (TCA) bottleneck, paired with a cross-attentive sequential decoder to extract rich, complementary features from raw RF data. At the decoder, we proposed a Cross-Attentive-Fusion (CAF) module that refines the displacement estimation at each stage of the sequential decoder.
Leveraging this architecture, MUSSE-Net employs a multi-stage refinement strategy, where each subsequent stage estimates residual displacements to iteratively refine strain outputs. Evaluated on both simulated RF and \textit{in vivo} datasets, USSE-Net and MUSSE-Net consistently outperforms existing models across key metrics, including SNR, contrast-to-noise ratio (CNR), and normalized root mean square error (NRMSE), demonstrating its robustness, precision, and effectiveness in real-world USE applications. The key contributions of this work are summarized as follows: 

\begin{itemize}
    \item We introduce MUSSE-Net, a residual-aware multi-stage unsupervised sequential framework for ultrasound strain elastography that progressively refines displacement estimation and achieves state-of-the-art performance on both simulated and \textit{in vivo} datasets.  

    \item We design a multi-stream encoder with Context-Aware Complementary Feature Fusion (CACFF) and a Tri-Cross Attention (TCA) bottleneck to effectively integrate contextual and modality-specific RF features for robust representation learning.  

    \item We introduce a sequential decoder with Cross-Attentive Fusion (CAF), which fuses multi-stream skip features through attention-guided refinement, enabling more accurate displacement estimation at each decoding level.  
\end{itemize}

\section{Datasets}
\label{subsection:dataset}
To evaluate the performance of the proposed method, we utilize three datasets from distinct sources: the Field II simulation dataset, the in vivo human dataset and a private clinical dataset. Detailed descriptions of each are provided below.

\subsection{Simulation dataset}
The publicly available simulation dataset developed by Tehrani and Rivaz for ultrasound strain elastography \cite{b22} is utilized. The dataset was generated using FIELD II simulations \cite{b32}, with Young’s modulus set between 18–23 kPa for inclusions and 40–60 $kPa$ for tissue, and a center frequency of 5 $MHz$. It consists of 24 phantoms (each with one or two inclusions at random positions), simulated at 10 strain levels (0.5–4.5\%) and 10 scatter positions. Since phantom 11 was unavailable, 23 phantoms were used: the first 19 for training and the last 4 for validation/testing. The validation set includes 36 sequences (instead of 40) as the last phantom had only 6 scatter positions.

\subsection{In Vivo dataset}

The open-access in vivo ultrasound elastography dataset from Delaunay \emph{et al.} \cite{b30} was used for training and evaluation. RF frames were acquired from a human arm using a Cicada 128PX system with a 7.5 MHz linear probe, yielding 310 sequences (19–127 frames each, 17,271 images total). Training data included sequences with lateral motion and decorrelation noise, while test sequences consistently contained at least one blood vessel. Temporal inputs were constructed from six consecutive frames, and model performance was assessed on 20 test samples (six frames each) from 13 acquisition sequences.

\subsection{Private BUET in vivo Breast Ultrasound dataset}
This clinical dataset was collected during 2012–2013 at the Medical Center of Bangladesh University of Engineering and Technology (BUET), Dhaka-1000, Bangladesh. In vivo breast ultrasound scans were acquired using a Sonix-Touch Research ultrasound system equipped with a linear L14-5/38 transducer. The transducer operated at a center frequency of 10 MHz with a sampling frequency of 40 MHz. RF sequence data were obtained from 23 subjects (ages 13–75 years) for training, while data from 5 subjects were reserved for testing and validation. In addition, one tissue mimicking phantom data acquired from this same machine has been used to evaluate the robustness of our proposed methods.

\begin{figure*}[!t]
    \centering
    \includegraphics[width=.9\textwidth, height=0.65\textheight, trim= 19cm 7cm 19cm 7cm, clip] {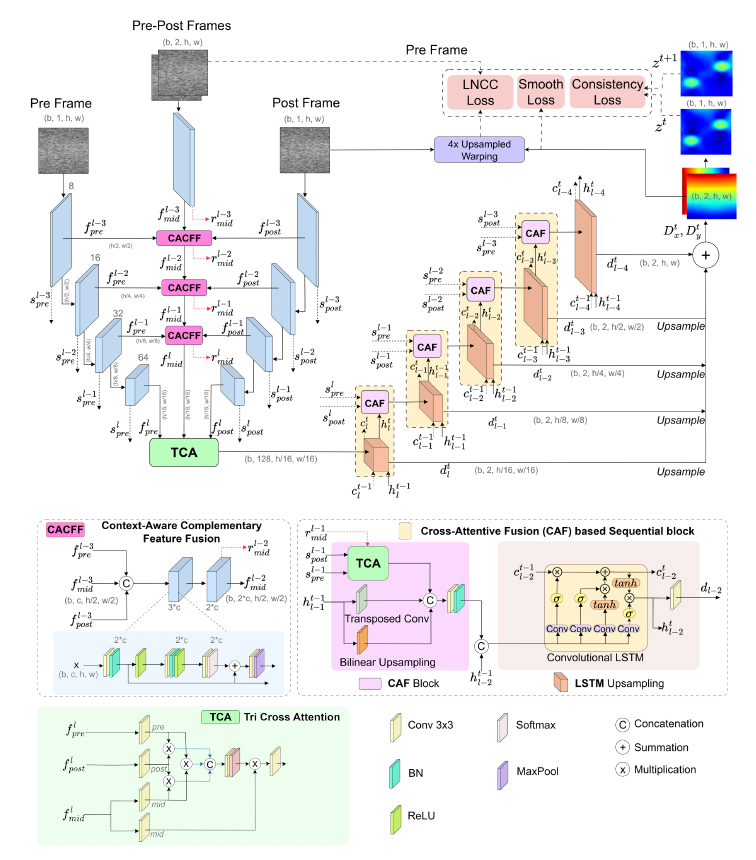}
    \caption{Architectural schematic of the proposed USSE-Net, which serves as the backbone of the MUSSE-Net framework.}
    \label{fig:5_model_arch}
\end{figure*}

\section{Proposed Method}
In this paper, we propose MUSSE-Net, a novel end-to-end residual-aware framework for strain elastography. Building on our backbone USSE-Net, and inspired in part by the sequential design of ReUSENet \cite{b30}, MUSSE-Net integrates a multi-stream fusion encoder–decoder with cross-attention. Its residual-aware design naturally extends to a multi-stage framework, where each stage refines strain estimation by modeling residual displacements. The architecture of the base network USSE-Net and the overall MUSSE-Net framework are described below.

\subsection{{USSE-Net Architecture}}
\label{subsection:basemodel}

Each stage of the propose MUSSE-Net is built upon a unified backbone architecture, USSE-Net, which processes RF frame sequences sequentially. 

The input consists of a reference frame $I_{pre} \in \mathbb{R}^{1 \times H \times W}$ and $T$ post-compression frames $I^{t}_{post} \in \mathbb{R}^{1 \times H \times W}$, where $t \in [1,2,\dots,T]$. For each pair $[I_{pre}, I^{t}_{post}]$, the model predicts displacement fields $D^t = (D^t_x, D^t_y) \in \mathbb{R}^{2 \times H \times W}$ where $D^t_y \in \mathbb{R}^{1 \times H \times W}$ is the axial displacement field and $D^t_x \in \mathbb{R}^{1 \times H \times W}$ corresponds to the lateral displacement field. Finally, the corresponding axial strain map $z^t$ is computed based on the estimated $D^t_y$. As shown in {Fig.~\ref{fig:5_model_arch}}, USSE-Net comprises three innovative key components: (i) a multi-stream encoder with Context-Aware Complementary Feature Fusion (CACFF), (ii) a Tri-Cross Attention (TCA) bottleneck, and (iii) a Cross-Attentive Fusion-based (CAF) Sequential decoder. Detailed descriptions and design rationales for each module are provided below.

\paragraph{\textbf{Context-Aware Complementary Feature Fusion-based Encoder:}}
\label{subsection:Encoder}
Conventional encoder designs for deep learning-based USE, whether single stream encoders that process concatenated pre- and post-compression frames $\left(I_{pre},, I^{t}_{post}\right)$ \cite{b30} or dual stream Siamese encoders \cite{b11}, often struggle to separate motion-specific complementary features from the shared contextual information present in both frames. To address this limitation, we introduce a Context-Aware Complementary Feature Fusion (CACFF) based encoder with three parallel branches (Fig.~\ref{fig:5_model_arch}). Two branches independently extract hierarchical high-level characteristics from $I_{pre}$ and $I^{t}_{post}$ using shared-weight residual downsampling modules. Each branch contains four residual downsampling blocks, with each block composed of two convolutional layers that generate features $f^{t,l}_{pre}$ and $f^{t,l}_{post}$ at layer $l$. The third branch has a similar structure but incorporates CACFF blocks that fuse the concatenated input $[I_{pre}, I^{t}_{post}]$ with $f^{t,l}_{pre}$ and $f^{t,l}_{post}$ as residuals. This design enables the learning of reciprocal and global contextual representations, producing fused features $f^{t,l}_{mid}$ across encoding levels except at the bottleneck. In this way, the three streams collectively capture motion-specific, reciprocal, complementary, and contextual features, which are then propagated to the bottleneck for attention-based fusion.

\paragraph{\textbf{Tri-Cross Attention-based Bottleneck:}}
\label{subsection:bottleneck}
At the bottleneck, the encoder outputs $f^{t,l}_{pre}$, $f^{t,l}_{post}$, and $f^{t,l}_{mid}$ are processed by the proposed Tri-Cross Attention (TCA) module. Unlike conventional correlation blocks that rely on local patch-wise matching with limited search ranges, TCA enables global feature interactions, capturing complex non-local dependencies between $I_{pre}$ and $I^{t}_{post}$. As illustrated in {Fig.~\ref{fig:5_model_arch}}, the TCA block first performs matrix multiplication across all paired combinations of $f^{t,l}_{pre}$, $f^{t,l}_{post}$, and $f^{t,l}_{mid}$ considering the paired features as key and query.
These multiplicative features are concatenated and passed through a convolutional layer to preserve the original channel dimension. A softmax activation layer is then applied to generate a probability distribution over the feature space that is the attention scores. The resulting attention map is finally dot-multiplied with $f^{t,l}_{mid}$ (value), enhancing critical global features. By computing region-level similarity across the entire spatial domain, TCA mitigates the limitations of cost-volume methods, reduces decorrelation noise and lateral artifacts, and improves the accuracy and structural coherence of strain estimation.

\paragraph{\textbf{Cross-Attentive Fusion based Sequential Decoder:}}
\label{subsection:decoder}
We propose a Cross-Attentive Fusion based Sequential Decoder to achieve accurate strain estimation with temporal coherence. The decoder integrates a ConvLSTM block \cite{b30} with our Cross Attention Fusion (CAF) mechanism. At each level $l$, the CAF block refines displacement estimates by fusing the previous decoder output $h^t_{l-1}$ ($t$ corresponds to the temporal state) with encoder skip features through attention-guided upsampling. Specifically, CAF generates an attention-weighted feature map from the skips, combines it with $h^t_{l-1}$ using learnable and bilinear upsampling, and forwards the result to the ConvLSTM. The ConvLSTM enforces temporal consistency by retaining hidden states $h^{t-1}_l$ across frame pairs and updating them to $h^t_l$ for sequential pre-compression frames $I_{pre}$ and post-compression frames $I^{t}_{post}, , t \in [1,2,3,\cdots,T]$, across multiple resolution levels. This temporal modeling enables the decoder to capture dynamic tissue deformation patterns effectively. The network outputs a cumulative displacement map, aggregated over all decoding levels, from which the axial strain map $z^t$ is derived using the Least Squares Strain Estimator (LSQSE). This design establishes a fully end-to-end framework for robust and temporally consistent strain estimation.

\subsection{MUSSE-Net Framework}
\label{subsection:two-stage}

\begin{figure}[t!]
\centerline{%
\includegraphics[height=.4\textheight, trim=18.5cm 5.5cm 18.5cm 5.5cm, clip, angle=0]{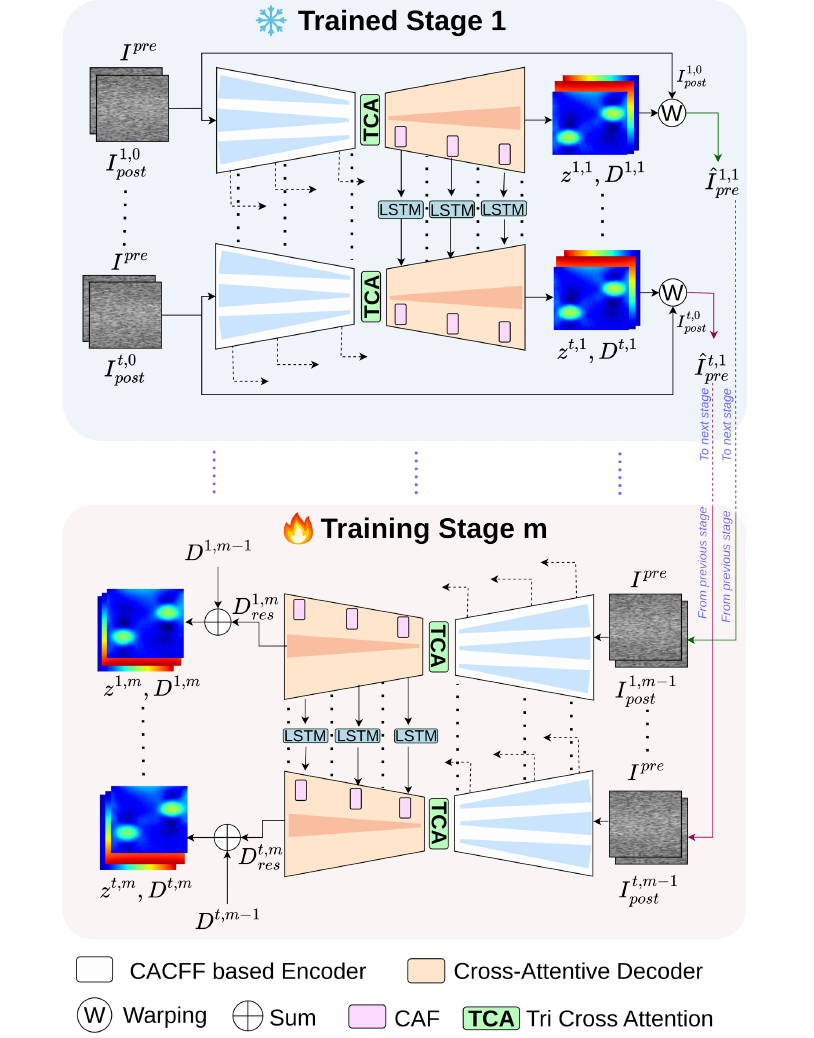}%
}
\caption{Block diagram of the proposed multi stage residual-aware framework, MUSSE-Net.}
\label{fig:two_stage}
\end{figure}

While USSE-Net serves as a powerful network for displacement and strain estimation, our ablation studies reveal that a single-stage network fails to fully capture the complexity of tissue motion, leaving scope for improving strain image quality. To address this, we propose MUSSE-Net, a residual-aware multi-stage framework that progressively refines strain estimation. By modeling residual displacements between the true pre-frame and the estimated pre-frames of earlier stages, MUSSE-Net produces high-quality strain maps with enhanced SNR in both target and background regions. An overview of MUSSE-Net is shown in {Fig.~\ref{fig:two_stage}}.  

At each stage $m$, the base USSE-Net ({Sec.~\ref{subsection:basemodel}}) takes as input a pair of RF frames: the true pre-frame $I_{pre}$ and a deformed post-frame $I^{t,m-1}_{post}$. It outputs displacement fields $D^{t,m} = \{D^{t,m}_{x}, D^{t,m}_{y}\}$, strain maps $z^{t,m}$, and warped post-frames $I^{t,m}_{post}$. For the first stage $(m=1)$, $I^{t,0}_{post}=I^{t}_{post}$; for later stages $(m>1)$, $I^{t,m-1}_{post}$ is set to the estimated pre-frame $\hat{I}^{t,m-1}_{pre}$. The warped post-frame at stage $m$ is given by
\begin{equation}
I^{t,m}_{post} = \textit{warp}(I^{t,m-1}_{post}, D^{t,m})
\label{eq:3_warp}
\end{equation}
where the warping function is defined as
\begin{equation}
I^{t,m}_{post} = I^{t,m-1}_{post}(x+D^{t,m}_{x}(x,y),\, y+D^{t,m}_{y}(x,y))
\label{eq:4_warping_function}
\end{equation}
To improve accuracy, we employ \emph{upsampled warping}: both $I^{t,m-1}_{post}$ and $D^{t,m}$ are first upsampled $(4\times)$ via bilinear interpolation, warped at this resolution, and then downsampled back to the original size.

During training of a particular residual stage ($m$), parameters of earlier stages remain frozen, and the network at stage $m$ learns to refine displacements by exploiting residual discrepancies between true and estimated pre-frames from immediate previous stage. Ideally, if the displacement $D^{t,m-1}$ is optimal, the warped pre-frame $I^{t,m-1}_{post}$ should match $I_{pre}$. To improve upon $D^{t,m-1}$, the estimated residual displacement is $D^{t,m}_{res}$. The refined displacement fields are updated by
\begin{equation}
D^{t,m} = D^{t,m-1} + D^{t,m}_{res}
\label{eq:final_disp}
\end{equation}
where, $D^{t,m}$ denotes the estimated displacements at stage $m$. Strain maps $z^{t,m}$ are then computed from $D^{t,m}$ using the LSQSE algorithm, making MUSSE-Net a fully end-to-end framework for USE. Although MUSSE-Net can be extended to $M$ stages, the optimal number $M_{opt}$ depends on the mean absolute difference between the estimated displacements of two consecutive stages. For our datasets, the mean absolute difference of displacements became negligible beyond two stages, and thus $M_{opt}=2$ was used in all experiments.

% \begin{figure}[t!]
% \centerline{\includegraphics[width=.\columnwidth, trim=18.5cm 5.5cm 18.5cm 5.5cm, clip]{figures/figure_3.pdf}}
% \caption{Block diagram of the proposed multi stage residual-aware framework, MUSSE-Net.}
% \label{fig:two_stage}
% \end{figure}

% \begin{figure*}[t!]
%     \centering
%     \includegraphics[width=\textwidth, height=.5\textheight, trim= 13.5cm 7cm 11.5cm 7cm, clip]{figures/figure_4_cb.pdf} %23_08
%     \caption{Qualitative comparison between the proposed framework and existing methods. From left to right, each column represents strain maps at varying strain levels.}
%     \label{fig:strain_results}
% \end{figure*}

\begin{figure*}[t!]
    \centering
    \includegraphics[width=\textwidth, trim= 13.5cm 7cm 13.5cm 4cm, clip]{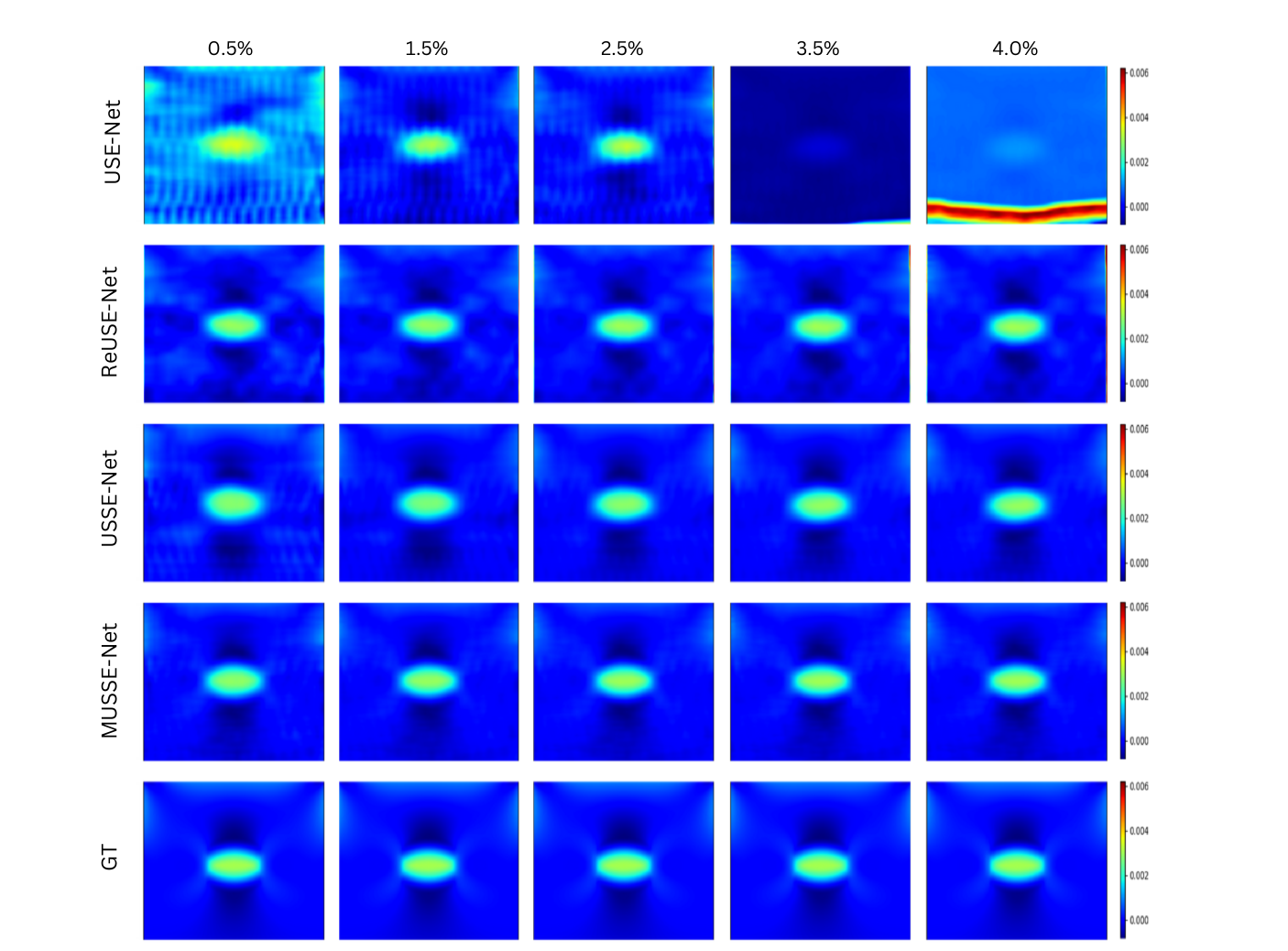} %23_08
    \caption{Qualitative comparison between the proposed framework and existing methods. From left to right, each column represents strain maps at varying strain levels.}
    \label{fig:strain_results}
\end{figure*}

\subsection{Loss Functions}

Embedding physical and anatomical priors into the loss formulation, our framework integrates three complementary loss components: 
\textit{similarity loss} ($L_{{sim}}$), which enforces alignment between pre-compression and deformation-compensated post-compression frames ($\hat I^t_{pre}$) via the predicted displacement field, ensuring physically plausible motion estimation; 
\textit{smoothness loss} ($L_{{smooth}}$), which promotes spatial regularity in the displacement field in line with soft tissue biomechanics; and 
\textit{consistency loss} ($L_{{con}}$), which preserves temporal coherence of predicted deformations across RF sequences.  
The component losses are defined as follows:  

\begin{itemize}
    \item \textit{Similarity Loss:}  
    At each temporal state $t$, the similarity loss $L^t_{sim}$ is derived from the Local Normalized Cross-Correlation (LNCC) index between the true pre-frame ($I_{pre}$) and the predicted pre-frame ($\hat I^t_{pre}$), obtained by warping the post-frame as described in {(\ref{eq:3_warp})}:
    \begin{equation}
    L^t_{sim} = 1 - LNCC,
    \label{eq:LNCC}
    \end{equation}
    where
    \begin{equation} 
    LNCC = \frac{1}{N} \sum_{i,j} 
    \frac{(W_1(i, j) - \mu_{W_1})(W_2(i, j) - \mu_{W_2})}{\sigma_{W_1}\sigma_{W_2}} 
    \end{equation}
    here, $N$ is the number of patches, $W_1(i,j)$ and $W_2(i,j)$ are pixel values within patches of $\hat I^t_{pre}$ and $I_{pre}$, while $\mu$ and $\sigma$ denote the patch-wise mean and standard deviation. The overall similarity loss is averaged across time:
    \begin{equation}
    L_{sim} = \frac{1}{T}\sum_{t=1}^{T} L^t_{sim}
    \label{eq:avg_LNCC}
    \end{equation}

    \item \textit{Consistency Loss:}  
    To enforce temporal coherence, consistency loss $L^t_{con}$ compares strain maps $z^t$ obtained from RF frame pairs $(I_{pre}, I^t_{post})$ with the preceding strain map $z^{t-1}$ \cite{b30}. Both maps are motion-compensated using their respective displacement fields:
    \begin{equation}
    L^t_{con} = LNCC(z^t, z^{t-1})
    \label{eq:L_con}
    \end{equation}
    and averaged across the sequence as
    \begin{equation}
    L_{con} = \frac{1}{T}\sum_{t=1}^{T} L^t_{con}
    \label{eq:avg_L_con}
    \end{equation}

    \item \textit{Smoothness Loss:}  
    Smoothness loss $L^t_{smooth}$ regularizes the estimated displacement field $D^t$, penalizing second-order gradients along axial and lateral directions to encourage spatial continuity:
    \begin{eqnarray}
    L^t_{smooth} &=& \sum_{i,j} \big( \, |\partial_x^2 D^t_{i,j}| \; + \; |\partial_x \partial_y D^t_{i,j}| \; + \; |\partial_y^2 D^t_{i,j}| \; \nonumber \\
    &&+ \; |\partial_y \partial_x D_{i,j}| \, \big)
    \label{eq:L_smoth}
    \end{eqnarray}
    then averaged across all temporal states:
    \begin{equation}
    L_{smooth} = \frac{1}{T}\sum_{t=1}^{T} L^t_{smooth}
    \label{eq:avg_L_smoth}
    \end{equation}
\end{itemize}

\noindent The overall training objective is thus formulated as
\begin{equation}
L_{total} = \alpha L_{sim} + \beta L_{con} + \gamma L_{smooth}
\label{eq:total loss}
\end{equation}
where the coupling factors are empirically set to $\alpha=1.0$, $\beta=0.2$, and $\gamma=0.3$.

\begin{figure*}[t]
    \centering
    \includegraphics[width=\textwidth, trim= 0cm 0cm 0cm 0cm, clip]{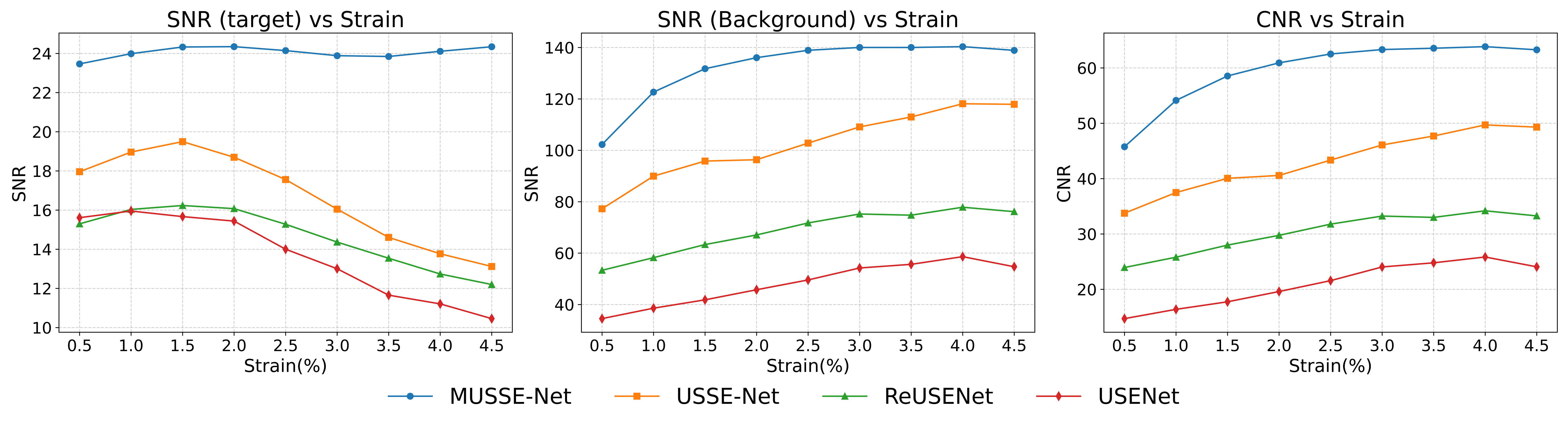}
    \caption{Analysis of average SNR (target and background) and CNR metrics at different stains. The leftmost shows $SNR_t$ vs strain, the middle one shows $SNR_{bg}$ vs strain, and the rightmost one depicts $CNR$ vs strain performance.}
    \label{fig:strain_metric_plot_results}
\end{figure*}

\begin{table*}[!t]
\centering
\setlength{\tabcolsep}{3pt}
% \resizebox{.8\textwidth}{!}{%
\begin{tabular}{lccccc}
\toprule
${Models}$ & ${SNR_{t}}$ & ${SNR_{bg}}$ & ${CNR}$ & ${NRMSE}$ & ${SNR_e}$ \\
\midrule
{USENet} & 13.66 ± 1.75 & 48.15 ± 7.27 & 20.98 ± 3.62 & 29.35 ± 0.77 & 4.43 ± 0.35 \\
\addlinespace
{ReUSENet} & 14.64 ± 2.23 & 68.43 ± 18.36 & 30.33 ± 8.06 & 2.03 ± 0.04 & 7.50 ± 0.58 \\
\addlinespace
{USSE-Net} & 16.69 ± 3.52 & 102.25 ± 33.36 & 43.11 ± 14.15 & 1.12 ± 0.12 & 9.16 ± 0.80 \\
\addlinespace
{MUSSE-Net} & 24.54 ± 3.66 & 132.76 ± 45.63 & 59.81 ± 20.38 & 1.31 ± 0.06 & 9.73 ± 1.08 \\
\bottomrule
\end{tabular}%
% }
\caption{Quantitative comparison of the proposed method with other approaches on the simulation dataset}
\label{table:qualitative_comparison}
\end{table*}

\subsection{Experimental Details}\label{subsection:training_details}

All evaluations were conducted using the PyTorch framework (version 2.5.1) on an NVIDIA V100 GPU (32GB). The batch size was set at 1 for each implementation. For training with simulation data, in each iteration of the training loop, the total number of post frames in an RF sequence was set to $T=9$ to form 9 pairs of RF images that can be fed to the models sequentially. For each pair, the first pre-compressed frame is selected as the reference ($I_{pre}$) and the post-compressed frames ($I^t_{post}$) are taken from the next $T=9$ frames of the sequence with linearly increasing strain. The training set to test set was $8.5:1.5$, exactly like the data split in \cite{b30} for valid comparison with these these methods. ADAM optimizer was used for gradient descent with an initial learning rate of $.001$ and $plateau$ as learning rate policy. All models were trained for $150$ epochs prior to evaluation. In this study, $M_{opt}$ was set to $2$, making MUSSE-Net a two-stage framework. The second stage of MUSSE-Net was trained for additional $100$ epochs under the same experimental conditions as the first stage of USSE-Net. After training, the evaluation was done on the other split of the data as done in \cite{b30}. Training with in vivo data and the private clinical data from BUET was performed in the exactly same experimental setting, except for the fact that in each iteration of the training loop, the number of post-compressed frames ($I^t_{post}$), $T$ was set to 5.

% The best epoch was selected by the validation loss criteria, the weights of the best epoch were saved, and the test dataset was evaluated on this best epoch model.

\subsection{Performance Evaluation Metrics}
\label{sec:perfom}
To evaluate the performance of all models, some very common and reliable metrics are used in the field of strain elastography estimation. The $SNR_t$, $CNR$, $SNR_e$ are used as the dominant metrics to compare performance. They are defined as,

\begin{equation}
{SNR}_t = \frac{\bar{s}_t}{\sigma_t},\quad 
{SNR}_e = \frac{\bar{s}}{\sigma}.
\label{eq:SNR}
\end{equation}

\begin{equation}
  {CNR} = \sqrt{\frac{2(\bar{s}_b - \bar{s}_t)^2}{\sigma_b^2 + \sigma_t^2}}.
  \label{eq:CNR}
\end{equation}

\noindent where \(\bar{s}_t\) and \(\bar{s}_b\) are the mean of the axial strain image at the target/lesion and background region, respectively. Similarly, \(\sigma_t\) and \(\sigma_b\) are the standard deviations of the lesion and background, respectively. The target and background regions are taken as elliptical regions inside the lesion and at the background, respectively. Here, \(\bar{s}\) and \(\sigma\) denote the general mean and standard deviation of the whole strain image that is used to calculate the $SNR_e$. We also calculated the normalized root-mean-square error (NRMSE) of the displacement field for the simulation test dataset, which is defined as
\begin{equation}
{NRMSE} = \frac{100 \times \sqrt{\frac{1}{N} \sum_{i=1}^{N} \left( \frac{w_{{GT},i} - w_{\theta,i}}{w_{{GT},i}} \right)^2}}{\sqrt{\frac{1}{N} \sum_{i=1}^{N} \left( \frac{w_{{GT},i}}{w_{{GT},i}} \right)^2}}.
\end{equation}
NRMSE score is calculated in percentage where \(w_{{GT}}\) and \(w_{\theta}\) are the ground truth and estimated axial displacement fields, respectively. We calculated the NRMSE of all the image pairs in the simulation test dataset and obtained the average value and standard deviation (SD). Note that all metrics are calculated in absolute values according to the equations not in dB.

\begin{figure*}[htbp]
    \centering
    % \hspace{1cm}
    \includegraphics[width=\textwidth, trim= 5cm 9.5cm 3cm 3cm, clip]{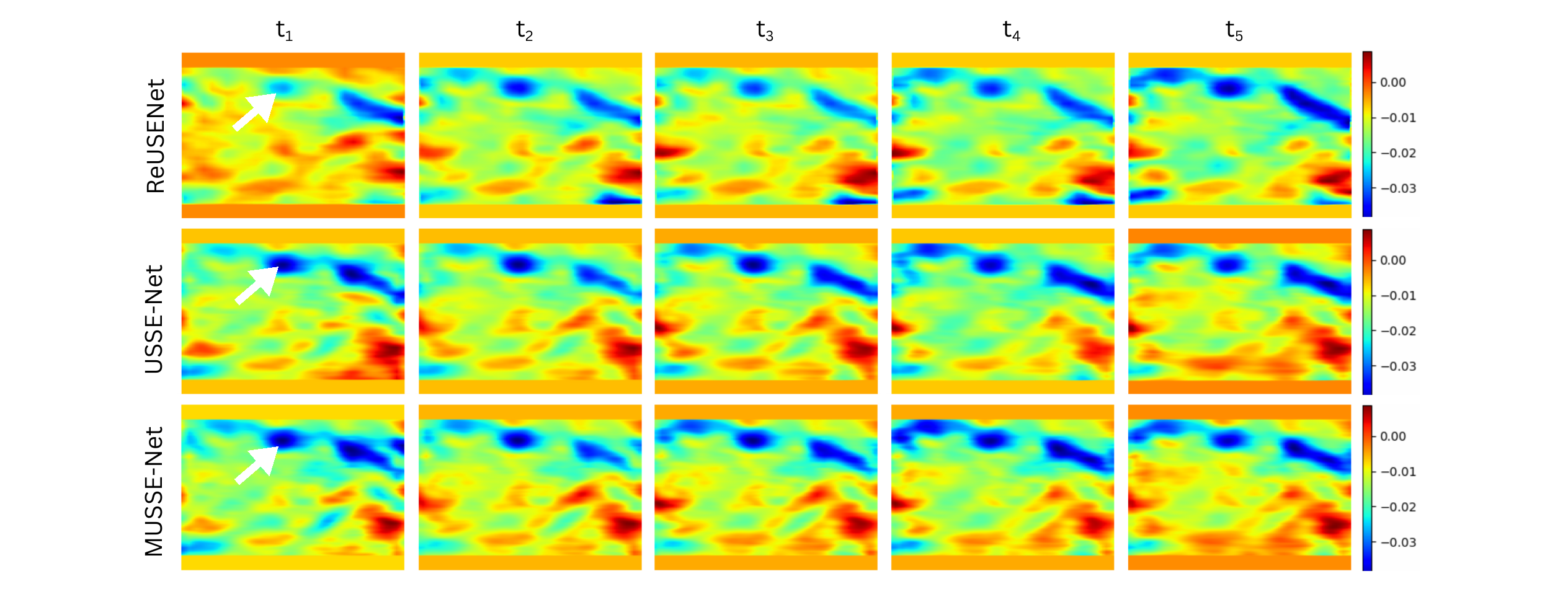}
    \caption{Qualitative strain results obtained from in vivo test data across a sequence of 5 ultrasound post-deformation RF frames. Each column shows the estimated strain map at that temporal frame by ReUSENet and out proposed USSE-Net and MUSSE-Net.}
    \label{fig:in vivo}
\end{figure*}

\begin{figure}[htbp]
    \centering
    \includegraphics[]{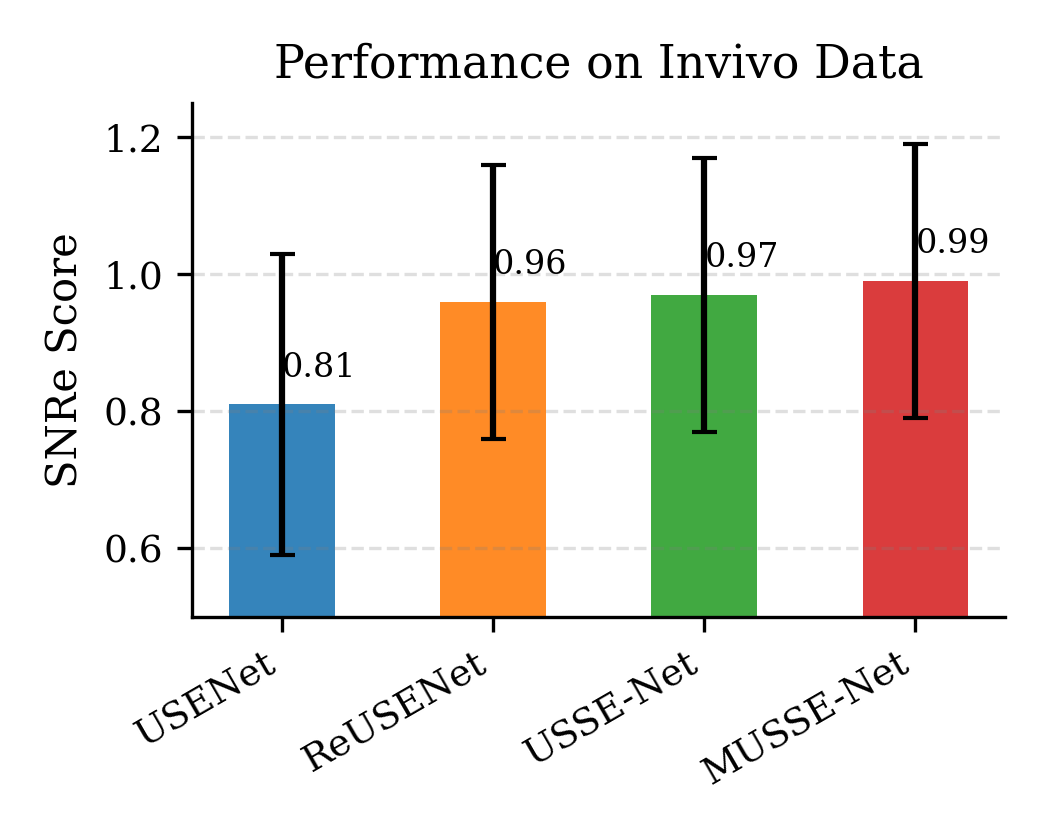}
    \caption{Mean elastographic SNR of different methods in the in vivo dataset.}
    \label{fig:in vivo_bar}
\end{figure}

\section{Results}
\subsection{Results on Simulation Dataset}

Quantitative and qualitative evaluations were conducted to assess the performance of the proposed MUSSE-Net and its backbone architecture, USSE-Net, using simulated ultrasound data. These models are compared against two state-of-the-art networks: USENet and ReUSENet \cite{b30}. The simulation dataset consisted exclusively of compressed RF frames, creating a challenging environment for displacement and strain estimation. To ensure a rigorous evaluation, a range of metrics were employed, capturing different aspects of image quality and estimation accuracy.

As shown in {Table~\ref{table:qualitative_comparison}}, both USSE-Net and MUSSE-Net outperform USENet and ReUSENet across all evaluation metrics. These metrics were computed per RF frame pair in the test dataset and averaged across all pairs. USSE-Net demonstrates strong improvements, and MUSSE-Net builds upon this by incorporating a residual-aware multi-stage framework, further enhancing estimation performance. For instance, USSE-Net achieves a target SNR ($SNR_t$) of 16.69, compared to 13.66 and 14.64 from USENet and ReUSENet, respectively. MUSSE-Net significantly boosts this further to 24.54, marking a 47.0\% improvement over USSE-Net ($p<$ 0.001). The performance gains are consistent across other metrics as well. The background SNR ($SNR_{bg}$) shows a 49.4\% increase over ReUSENet, reaching 102.25 in USSE-Net versus 68.43 ($p <$ 0.001). MUSSE-Net enhances the background as the ($SNR_{bg}$) is further increased to 132.76. Similarly, USSE-Net achieves a CNR of 43.11, a 42.1\% improvement over ReUSENet ($p<$ 0.001) which achieves further improvement in MUSSE-Net reaching a CNR of 59.81. The elastographic SNR ($SNR_e$) increases from 7.50 in ReUSENet to 9.16 in USSE-Net and further to 9.73 in MUSSE-Net—a 29.7\% increase. Additionally, the NRMSE is substantially reduced by 44.5\% in USSE-Net compared to ReUSENet, indicating higher accuracy in displacement estimation. While MUSSE-Net shows a slightly higher NRMSE than USSE-Net (1.31 vs. 1.12), this tradeoff is acceptable due to the added residual displacement refinement, which introduces minor additive noise but significantly boosts clinically relevant metrics like SNR and CNR.

The qualitative results, illustrated in {Fig.~\ref{fig:strain_results}}, provide further evidence of MUSSE-Net’s superiority. Strain maps were generated across a range of simulated strain levels (0.5\% to 4.5\%). USENet failed to produce reliable maps at higher strain levels, while ReUSENet, though more robust due to its convolutional LSTM decoder, suffered from noisy outputs and blurred lesion boundaries. In contrast, both USSE-Net and MUSSE-Net produced cleaner and smoother axial strain maps. Notably, MUSSE-Net preserved lesion edges more effectively, reduced lateral decorrelation noise, and enhanced overall map quality, regardless of strain level. Further analysis of strain consistency is presented in {Fig.~\ref{fig:strain_metric_plot_results}}, which plots the mean values of $SNR_t$, $SNR_{bg}$, and CNR across varying strain levels. USENet and ReUSENet exhibit significant performance degradation at higher strains, with metrics sharply declining as tissue deformation increases. USSE-Net mitigates this trend to some extent but still shows reduced stability under large strains. In contrast, MUSSE-Net, with it's residual-aware framework, maintains highly consistent performance across all strain levels, demonstrating its robustness and adaptability. Even at higher strains, the proposed model maintains high $SNR_t$ and $SNR_{bg}$, indicating reliable target and background quality under challenging conditions.

Therefore, it can be clearly observed that both USSE-Net and MUSSE-Net demonstrate substantial improvements over existing methods in displacement and strain estimation tasks.

\begin{figure*}[t]
    % \hspace{-0.55cm}
    \includegraphics[width=\textwidth, trim= 3cm 10cm 5cm 3.5cm, clip]{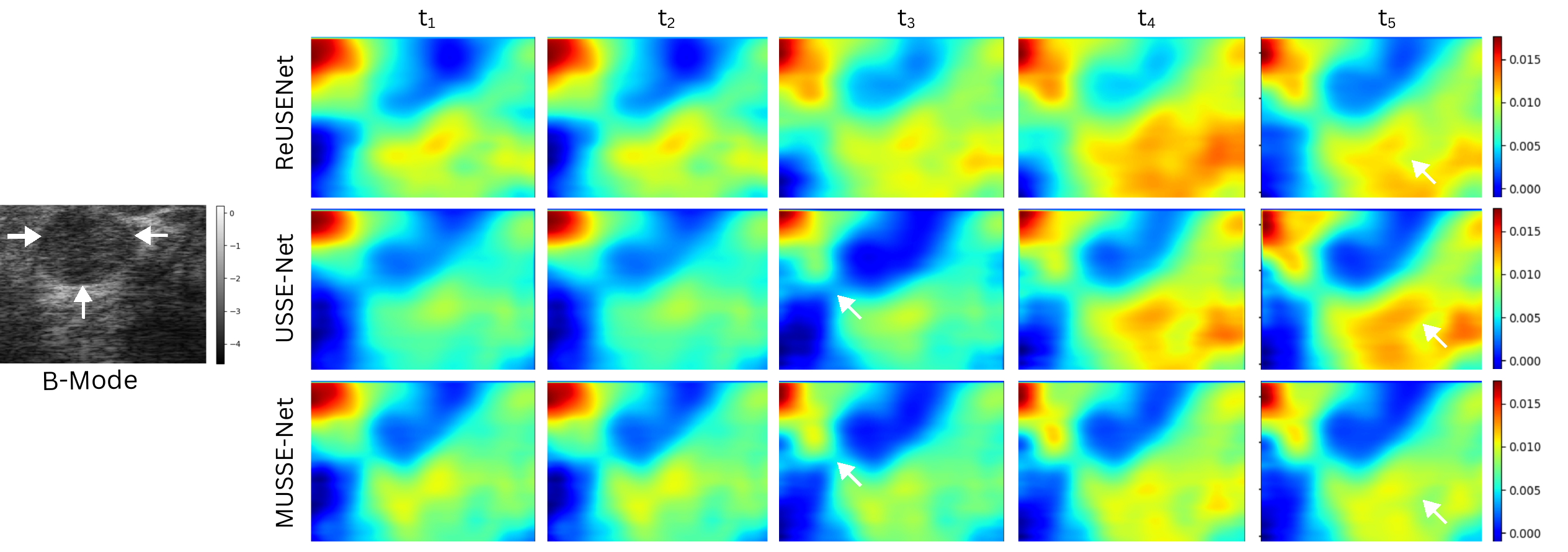}
    \caption{Comparative qualitative strain results on the private BUET in vivo breast ultrasound dataset using ReUSENet, and our proposed USSE-Net and MUSSE-Net.}
    \label{fig:clinical}
\end{figure*}

\subsection{Results on \textit{in vivo} Dataset}
The evaluation on the in vivo dataset further validates the effectiveness of the proposed USSE-Net and MUSSE-Net architectures. As illustrated in {Fig.~\ref{fig:in vivo_bar}}, their performance is compared against established methods such as USENet and ReUSENet. For this dataset, the $SNR_e$ is used as the primary evaluation metric, reflecting the quality and consistency of the estimated strain maps. Among all models, USENet exhibits the poorest performance on the in vivo data, achieving a mean $SNR_e$ of only 0.81. ReUSENet, with its sequential decoder design, performs noticeably better, reaching a mean $SNR_e$ of 0.96. The proposed USSE-Net offers a modest yet meaningful improvement, achieving a mean $SNR_e$ of 0.97, indicating a more stable and accurate strain estimation.

However, the proposed MUSSE-Net framework shows an improvement of 3.125\% over ReUSENet, achieving a mean $SNR_e$ of 0.99. These gains are further corroborated by the visual results shown in {Fig.~\ref{fig:in vivo}}, which presents strain maps for a different in vivo test sample. As can be seen, MUSSE-Net produces noticeably clearer and more anatomically accurate strain maps, particularly in delineating blood vessel regions. On the other hand, ReUSENet fails to capture the target region at time step $t_1$, while both USSE-Net and MUSSE-Net successfully identify and track the blood vessel throughout the temporal sequence. The proposed methods also yield more consistent and well-localized strain estimations over time. These results on the in vivo dataset highlight the robustness and clinical potential of the proposed methods, particularly MUSSE-Net, in producing high-quality, temporally consistent strain maps under real-world imaging conditions.

\subsection{Results on Private BUET \textit{in vivo} Breast Ultrasound Dataset}
{Fig.~\ref{fig:clinical}} shows qualitative strain results on the private BUET in vivo breast ultrasound test set for ReUSENet, USSE-Net, and MUSSE-Net. The test case contains a real lesion clearly visible in the B-mode image. ReUSENet fails to generate a consistent strain field and does not properly capture the lesion shape, particularly at the $3^{rd}$ and $4^{th}$ temporal RF frames, highlighting the limitations of its architecture. In contrast, both USSE-Net and MUSSE-Net consistently reconstruct the strain field and preserve lesion morphology across the entire RF sequence. USSE-Net produces better strain maps around the lesion (notably in the $2^{nd}$ and $3^{rd}$ frames), but also introduces lateral background noise. MUSSE-Net, supported by the residual-aware framework, effectively reduces this noise while maintaining robust and accurate strain estimation throughout the sequence, especially in the final RF frames.

\begin{figure}[htbp]
    \centering
    \includegraphics[width=\columnwidth, height=.4\columnwidth, trim= 10cm 12cm 11cm 7cm, clip]{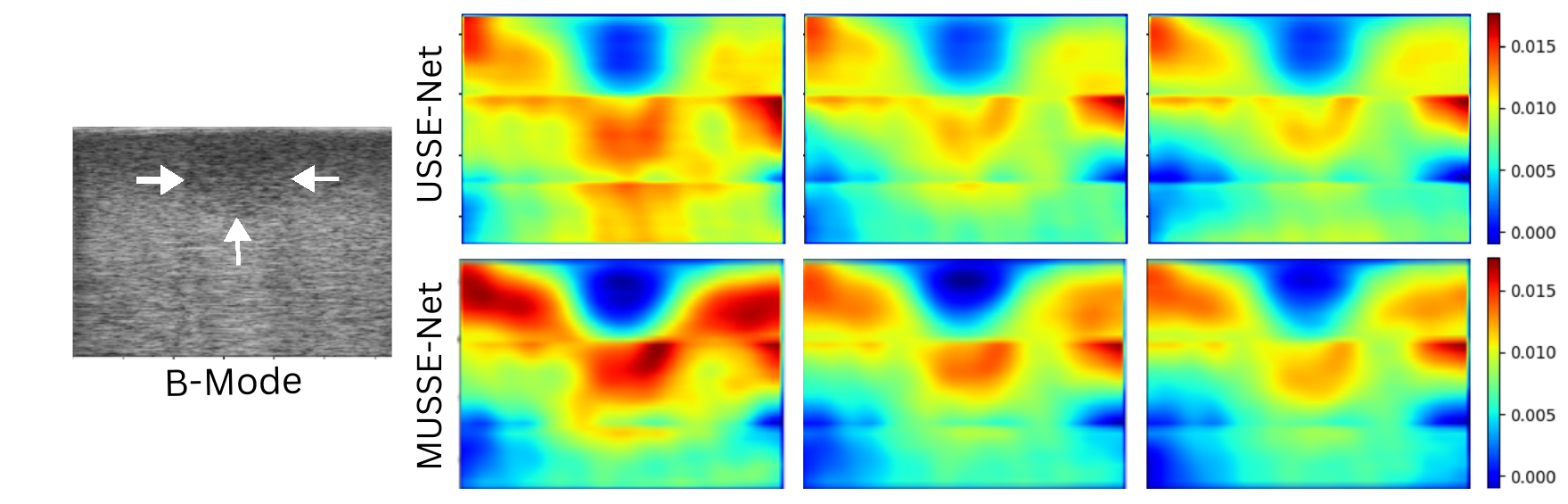}
    \caption{Qualitative strain results of the proposed methods evaluated on a phantom sample of the private BUET breast ultrasound dataset after being trained on the BUET in vivo dataset.}
    \label{fig:phantom}
\end{figure}

{Fig.~\ref{fig:phantom}} shows the qualitative strain outputs of USSE-Net and MUSSE-Net, trained on the BUET in vivo breast ultrasound dataset and evaluated on a tissue mimicking phantom sample excluded from training. Both networks generalize well to this unseen and out-of-domain case, producing reliable strain estimates that demonstrate the robustness, transferability, and domain gap generalization of the proposed framework.

\section{Discussion}
\subsection{Ablation Study}
To systematically validate the effectiveness of our proposed MUSSE-Net, we conducted a comprehensive ablation study. Starting from the baseline ReUSENet, we incrementally introduced key architectural design components, including the CACFF-based encoder, TCA-based bottleneck, CAF-based sequential decoder, and finally, the multi-stage design of MUSSE-Net. The quantitative results of this progressive model evolution are summarized in {Table~\ref{table:ablation}}.

\begin{figure*}[t]
    \centering
    \includegraphics[width=\textwidth, trim= 13cm 15.5cm 17cm 4cm, clip]{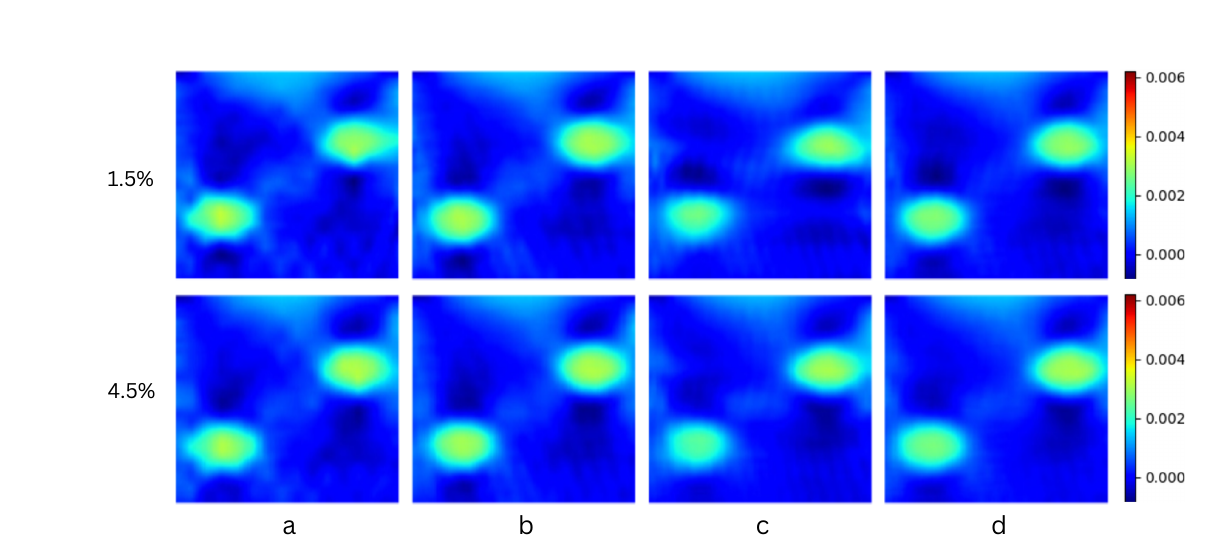}
    \caption{Qualitative results of the ablation study on simulation data. (a) ReUSENet, (b) Our CACFF based encoder with TCA bottleneck and ReUSENet Decoder, (c) USSE-Net, (d) MUSSE-Net.}
    \label{fig:ablation}
\end{figure*}

\begin{table*}[htbp]
\centering
% \resizebox{.75\textwidth}{!}{
\begin{tabular}{lccccc}
% \toprule
\toprule
\multicolumn{1}{l}{${Model}$} & ${SNR_{t}}$ & ${SNR_{bg}}$ & ${CNR}$    & ${SNR_e}$ \\
\midrule
$a)$ Baseline & 14.64 ± 2.23   & 68.43 ± 18.36 & 30.33 ± 8.06  & 7.50 ± 0.58  \\
$b)$ a + CACFF + TCA &  15.45 ± 2.49 & 98.36 ± 31.58 & 42.12 ± 13.07 & 8.85 ± 0.96  \\
$c)$ USSE-Net (b + CAF Decoder) & 16.69 ± 3.52 & 102.25 ± 33.36 & 43.11 ± 14.15  & 9.16 ± 0.80  \\
$d)$ MUSSE-Net & 24.54 ± 3.66 & 132.76 ± 45.63 & 59.81 ± 20.38  & 9.73 ± 1.08  \\
\bottomrule
% \bottomrule
\end{tabular}
% }
\caption{Ablation results showing the importance of each block in strain image reconstruction; a) Baseline which is ReUSENet, b) Our proposed CACFF + TCA encoder with ReUSENet Decoder, c) Our proposed network USSE-Net and d) The proposed framework MUSSE-Net}
\label{table:ablation}

\end{table*}

We first focused on the encoder and bottleneck components. By integrating the proposed CACFF encoder with the TCA-based bottleneck (a), we addressed the inherent limitations of single- and dual-stream encoder architectures. The multi-stream CACFF encoder effectively disentangles structural and motion features, while the TCA bottleneck mitigates the drawbacks of conventional convolutional or cross-correlation-based bottlenecks by enhancing feature extraction through the attention mechanism. The impact of these components is immediately evident: compared to the baseline ReUSENet, the target SNR ($SNR_t$) increased by 0.81, reaching 15.45. Notably, the background SNR ($SNR_{bg}$) showed a substantial 44\% improvement, rising from 68.43 to 98.36, and $SNR_e$ improved by 18\%, from 7.50 to 8.85. These gains are visually supported in {Fig.~\ref{fig:ablation}}, where smoother and more detailed strain maps clearly reflect the improved estimation. This demonstrates the effectiveness of CACFF in extracting context-aware complementary features and the TCA mechanism in reducing lateral decorrelation noise.

In the next step of our ablation study (b), we replaced the original ReUSENet decoder with the proposed CAF-integrated sequential decoder, resulting in our backbone model, USSE-Net. This new decoder applies cross-attention fusion between skip connections and temporally encoded decoder features, refining displacement estimation at each decoding stage. As shown in {Table~\ref{table:ablation}}, this modification led to consistent improvements across all evaluation metrics. The target SNR ($SNR_t$) increased to 16.69, while $SNR_e$ rose to 9.16, indicating enhanced strain map quality. Additionally, both $SNR_{bg}$ and CNR improved significantly, confirming the decoder’s ability to effectively fuse spatial and temporal information for accurate displacement refinement. Finally, we introduced the full residual-aware multi-stage framework detailed in Section \ref{subsection:two-stage}, resulting in the final version of MUSSE-Net. This design stacks multiple stages (two stages in our case) of USSE-Net, where the residual between the predicted and reference displacements is iteratively refined. As shown in row (d) of {Table~\ref{table:ablation}}, this multi-stage implementation achieves significant performance gains across all metrics. $SNR_t$ jumps to 24.54, marking a 47\% increase over USSE-Net, while $SNR_e$ reaches 9.73, further improving the quality of the elastographic output. Background SNR and CNR also show substantial enhancements, rising to 132.76 and 59.81, respectively. These results confirm that the multi-stage refinement not only resolves the limitations of single-stage estimation but also amplifies clinically important quality measures such as target contrast and background clarity.

The qualitative improvements brought by MUSSE-Net are illustrated in {Fig.~\ref{fig:ablation}}, specifically columns (c) and (d). Compared to the first-stage output (USSE-Net) and the earlier models (rows a and b), MUSSE-Net significantly reduces lateral decorrelation artifacts and produces smoother, more anatomically accurate strain maps without any additional post denoising methods. The effectiveness of the multi-stage approach is particularly evident under varying deformation levels. As shown in {Fig.~\ref{fig:ablation}}, the first row displays performance at low strain (1\%), while the second row corresponds to high strain (4.5\%). Across both conditions, MUSSE-Net consistently maintains image quality and structural integrity, validating its robustness in diverse tissue deformation scenarios. Therefore, our comprehensive ablation study demonstrates the progressive contributions of each architectural component, culminating in a highly effective multi-stage framework. MUSSE-Net not only surpasses existing models in both quantitative metrics and visual fidelity but also establishes a strong foundation for robust, unsupervised strain estimation in ultrasound elastography.

\subsection{Limitations and Future Works}

As demonstrated in the results and ablation study sections, MUSSE-Net exhibits strong performance; however, it is not without limitations. The current multi-stage implementation requires substantially longer training times and incurs higher inference latency compared to USSE-Net. Addressing these challenges in future work, we plan to explore lightweight yet effective backbone architectures and pruning strategies to reduce computational complexity, alongside large-scale validation on diverse in vivo datasets. Additionally, the sequential nature of the network constrains batch sizes to one, further contributing to computational overhead. Moving forward, we also aim to design more efficient mechanisms for incorporating time-domain features in a scalable manner.

\section{Conclusion}
In this study, we introduced MUSSE-Net, a residual-aware unsupervised multi-stage deep learning framework designed to enhance consistency and fidelity in strain elastography. Building upon the base architecture USSE-Net, MUSSE-Net integrates a CACFF encoder for effective feature disentanglement, a TCA bottleneck for robust correspondence matching, and a CAF-ConvLSTM decoder to enforce temporal coherence. This combination enables substantial gains in both foreground and background SNR, robust suppression of decorrelation artifacts, and state-of-the-art quantitative performance. Qualitative evaluations further highlight improved target delineation and significant reduction of lateral noise artifacts. Unlike single-stage methods, the proposed multi-stage refinement strategy proves critical for capturing the complete displacement field and generating high-quality strain maps. A key strength of our work lies in the extensive evaluation on in vivo breast ultrasound datasets, particularly the private BUET dataset, which provides a rare and challenging benchmark for validating model generalizability. MUSSE-Net consistently demonstrates noticeable improvements on these in vivo cases, underscoring its robustness and clinical potential. Overall, this work marks an important step toward advancing unsupervised strain elastography and bridging the gap toward real-world clinical adoption.

{
    \small
    \bibliographystyle{ieeenat_fullname}
    \bibliography{main}
}

% WARNING: do not forget to delete the supplementary pages from your submission 
% \input{sec/X_suppl}
% \bibliographystyle{unsrtnat}
% \bibliography{references}
\end{document}